\documentclass[11pt]{article}

\usepackage[preprint]{acl}

\usepackage{times}
\usepackage{latexsym}
\usepackage{enumitem}
\usepackage[T1]{fontenc}

\usepackage[utf8]{inputenc}

\usepackage{microtype}

\usepackage{inconsolata}

\usepackage{graphicx}
\usepackage{multirow}
\usepackage{booktabs}
\usepackage{amssymb}
\usepackage{amsmath}
\usepackage{xcolor}
\usepackage{bbding}
\usepackage{cuted}
\usepackage{tcolorbox}
\usepackage{caption}
\tcbuselibrary{breakable}
\usepackage{fontawesome}
\usepackage{listings}

\usepackage{algorithm}
\usepackage{algorithmic}

\tcbset{breakable}

\definecolor{c1}{HTML}{344C11}
\definecolor{c2}{HTML}{7e0f12}

\newcommand{\data}{QUASAR}
\title{Symbolic and Abstractive Reasoning with Complex Visual Queries}

\author{
    Yichi Zhang$^\spadesuit$\footnotemark[1],
    Jingdian Lu$^\spadesuit$\thanks{$\quad$ Equal Contribution.},
    Zhuo Chen$^\spadesuit$,
    Lingbing Guo$^\diamondsuit$,
    Jun Xu, $^\clubsuit$ \\
    \textbf{Wen Zhang}$^\spadesuit$,
    \textbf{Huajun Chen}$^\spadesuit$\\
    $^\spadesuit$ Zhejiang University,
    $^\diamondsuit$ Nanjing University,
    $^\clubsuit$ Ant Group\\
    \texttt{
    \{zhangyichi.each,zhang.wen,huajunsir\}@zju.edu.cn 
    }\\
}

\begin{document}
\maketitle
\begin{abstract}
Understanding and reasoning over abstract visual content remains a challenge for current multi-modal large language models (MLLMs). In this paper, we explore a novel abstract data type termed complex visual query (CVQ), designed to probe symbolic and abstractive reasoning, which is a critical yet underexplored dimension of human-like neuro-symbolic reasoning for MLLMs. We present a comprehensive investigation from three perspectives: \textbf{Data $\times$ Paradigm $\times$ Exploration}. Specifically, we propose a scalable pipeline for synthesizing CVQs grounded in large-scale multi-modal knowledge graphs, generating a diverse dataset encompassing 14 distinct query types via systematic combinations of first-order logic operators. We further introduce a two-stage training framework that progressively equips MLLMs with robust visual reasoning capabilities. We conduct extensive experiments to rigorously evaluate MLLMs across multiple dimensions, including reasoning performance on CVQs, as well as cross-task and cross-scenario generalization. We believe our work opens new perspectives and avenues for advancing the reasoning frontiers of MLLMs.
\end{abstract}

\section{Introduction}
\begin{figure}[t]
  \centering
  \includegraphics[width=0.5\textwidth]{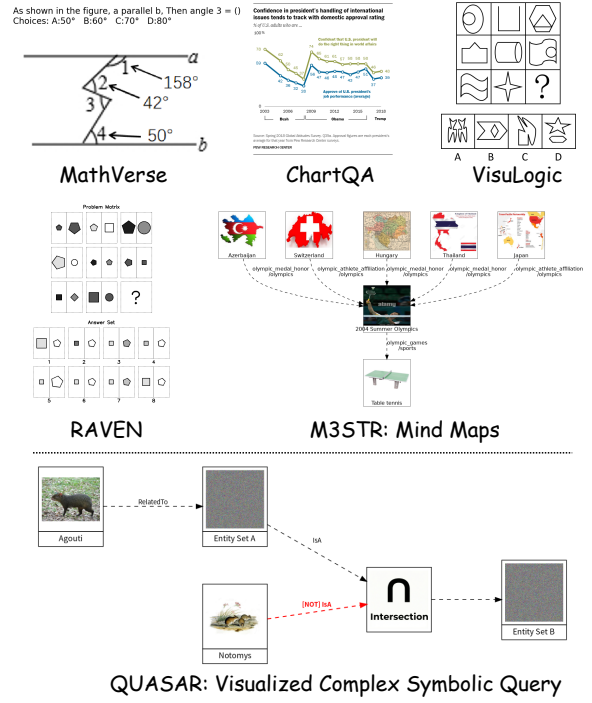}
  \caption{Data sample from different abstract reasoning benchmarks compared with our {\data} data.}
  \label{figure::introduction}
  \vspace{-16pt}
\end{figure}

Although multi-modal large language models (MLLMs)~\cite{MLLM-survey} have made remarkable strides in visual comprehension and reasoning, understanding and reasoning over abstract visual information remains a fundamental challenge. Visual information manifests in diverse forms, e.g., statistical charts~\cite{ChartQA}, geometric figures~\cite{Math-VISTA}, and mind maps~\cite{m3str}, often encoding rich \textbf{human-defined implicit semantic information} that distinguishes such abstract representations from natural, concrete visual content. As illustrated in Figure~\ref{figure::introduction}, we present representative samples from several abstract reasoning benchmarks. Enabling MLLMs to understand such information and perform deep, structured reasoning over it has emerged as a central focus of the research community.

\begin{figure*}[h]
  \centering
  \includegraphics[width=\textwidth]{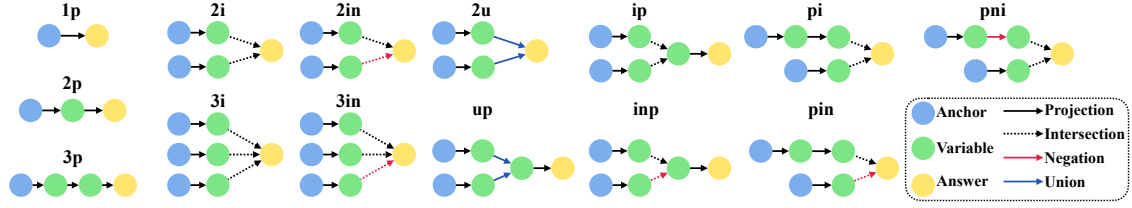}
  \caption{14 different complex query types in our {\data} data based on projection/intersection/union/negation.}
  \label{figure::query_type}
  \vspace{-16pt}
\end{figure*}
\par Research on abstract visual representations spans diverse focuses: studies on statistical charts probe MLLMs' data analysis capabilities and sensitivity to statistical measures, research on geometric figures emphasizes theorem proving, and studies on mind maps~\cite{STAR-KGRPO} explore how models can reason over multi-modal information in a structured, human-like manner. In the realm of mind maps, existing work typically samples data from multi-modal knowledge graphs (MMKGs) and visualizes it as graph-structured images to serve as reasoning references, which is largely confined to single-hop reasoning. Such an approach fails to fully exploit the rich compositional structures and complex reasoning patterns inherent in MMKGs. In fact, by combining diverse entities, relations, and first-order logic operators~\cite{query2box}, one can construct \textit{complex queries} that enable neuro-symbolic reasoning far beyond simple semantic networks. Casting such complex queries in visual form yields a novel and challenging data type for MLLMs: \textbf{complex visual queries (CVQs)}.

\par To address this research gap, we systematically investigate symbolic and abstractive reasoning with CVQs. Our work targets three core challenges in this emerging field:
\begin{itemize}[leftmargin=8pt]
    \item \textbf{C1: Data Scarcity.} How can we construct diverse and scalable CVQ datasets for research?
    \item \textbf{C2: Paradigm Design.} How can MLLMs effectively acquire and solve CVQs, and how should CVQ performance be reliably evaluated?
    \item \textbf{C3: Generalization.} Do the CVQ reasoning capabilities of MLLMs generalize to out-of-distribution scenarios?
\end{itemize}
In response to the three challenges outlined above, we makes 3 core contributions. To address \textbf{C1}, we design and implement a CVQ data synthesis engine that samples diverse query patterns from large-scale MMKG datasets and renders them as structured visual representations. The resulting data forms a new benchmark, \textbf{\textsc{\data}} (\underline{\textbf{V}}isual \underline{\textbf{Qu}}ery \underline{\textbf{a}}s \underline{\textbf{S}}ymbolic \underline{\textbf{A}}nd \underline{\textbf{R}}easoning), covering 14 canonical first-order logic (FOL) query combinations~\cite{query2box} along with comprehensive, fine-grained chain-of-thought annotations.
 
\par To address \textbf{C2}, we formulate two novel tasks: \textbf{\textit{Complex Query Understanding} (CQU)} and \textbf{\textit{Complex Query Reasoning} (CQR)}, which are tailored to the characteristics of MLLMs, with corresponding dataset splits, a two-stage training framework, and well-defined evaluation metrics, forming a coherent training-evaluation paradigm for CVQ research. To address \textbf{C3}, we conduct extensive experiments to assess the performance of current MLLMs on CVQ tasks, their generalizability across diverse query types, and the transferability of CVQ capabilities to out-of-distribution visual reasoning tasks. 
\par Our overarching goal is not merely to teach MLLMs a set of specific new tasks, but to cultivate new reasoning patterns that enhance their generalization capacity across diverse visual inputs. Grounded in a three-dimensional framework of \textbf{Data $\times$ Paradigm $\times$ Exploration}, we provides a panoramic view of CVQ in MLLMs, charting a promising new direction for future research.

\section{Related Works}
\begin{figure*}[h]
  \centering
  \includegraphics[width=\textwidth]{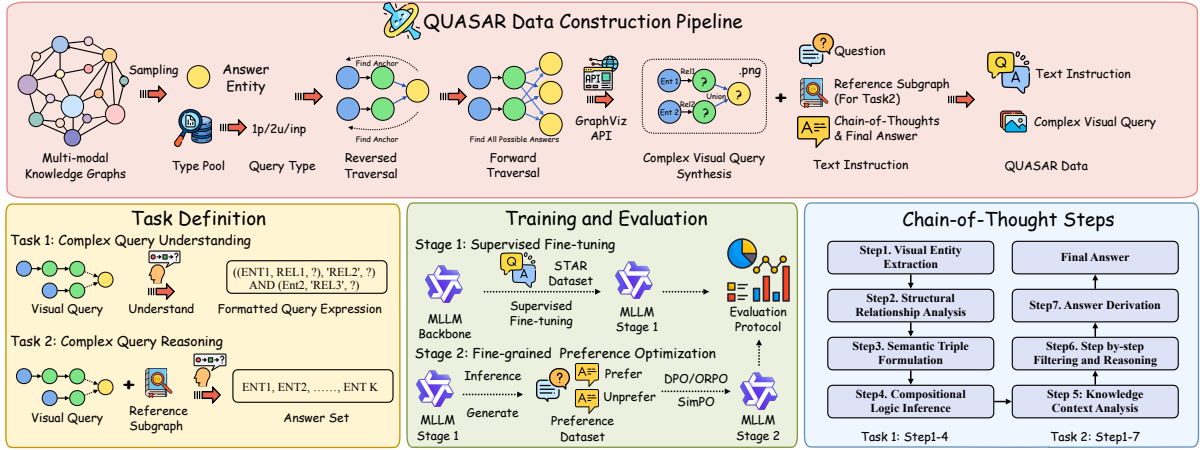}
  \caption{Overview of the {\data}'s construction pipeline. We present the detailed steps, definition of the CQU/CQR task, training pipelines, and CoT prompt design of our work.}
  \label{figure::overview}
  \vspace{-16pt}
\end{figure*}
\noindent \textbf{Abstract Visual Reasoning.} Understanding and reasoning over concrete visual information~\cite{MLLM-survey} have been extensively studied. However, reasoning over abstract visual representations such as charts~\cite{ChartQA}, puzzles~\cite{RAVEN}, and mind maps~\cite{m3str} remains in its infancy, as they carry unique structural patterns that pose greater challenges for higher-order reasoning. M3STR~\cite{m3str} and STAR~\cite{STAR-KGRPO} focus on synthesizing multi-modal mind maps to support structured visual reasoning. RAVEN~\cite{RAVEN} and Multi-STAR~\cite{MultiStar} generate abstract geometric patterns for MLLM evaluation. GITA~\cite{GITA} and DynamicGTR~\cite{DynamicGTR} provide benchmarks with accompanying training methods that enable MLLMs to tackle graph theory problems from a visual perspective. We extend this line of research to complex query reasoning~\cite{BetaE} in the visual modality, which has not been previously explored.

\noindent \textbf{Complex Query Answering (CQA) in KGs.} Traditional KG reasoning~\cite{DBLP:conf/nips/TransE} is largely confined to single-hop queries involving only projection operations. CQA extends this by incorporating first-order logic operators such as union, intersection, and negation, enabling more expressive and compositional reasoning patterns. Existing approaches~\cite{BetaE,kgTransformer} predominantly employ embedding-based models that encode complex queries as dense vector representations. More recently, several works~\cite{LARK,LACT} have explored the potential of LLMs to tackle CQA as a text-based reasoning task.

\section{Task Definition}
We focus on complex queries in KGs, which can be denoted as $\mathcal{KG}=(\mathcal{E}, \mathcal{R}, \mathcal{T})$, where $\mathcal{E}, \mathcal{R}$ are the entity and relation sets. $\mathcal{T} \subseteq \mathcal{E} \times \mathcal{R} \times \mathcal{E}$ denotes the triple set. A complex query $q(V_?)$ with target variable $V_?$ is characterized by a set of anchor entities $\mathcal{V}_a \subset \mathcal{E}$ serving as known starting points, and a set of intermediate variables $\mathcal{V}_e = \{V_1, \ldots, V_k\}$ representing unknown intermediate entities. Following the standard formulation~\cite{query2box}, a complex query is expressed in disjunctive normal form (DNF) as:
\begin{equation}
    q(V_?)=V_?. \exists V_1, \cdots, V_k: c_1 \vee c_2 \vee ... \vee c_n
\end{equation}
where $c_i$ represents a conjunctive ($\wedge$) query as:
\begin{equation}
    c_i = e_{i,1} \wedge e_{i,2} \wedge \dots \wedge e_{i,j}
\end{equation}
Each atom $e_{i,j}$ is an atomic formula or its negation, i.e., $e_{i,j} = r(V_i, V_j)$ or $e_{i,j} = \neg r(V_i, V_j)$, where $r \in \mathcal{R}$ denotes a relation projection from entity variable $V_i$ to $V_j$, and $V_i, V_j \in \mathcal{V}_a \cup \mathcal{V}_e$. The four fundamental first-order logic (FOL) operations underlying complex queries are \textbf{projection} (p), \textbf{intersection} (i), \textbf{union} (u), and \textbf{negation} (n). As illustrated in Figure~\ref{figure::query_type}, we investigate 14 canonical query types (1p, 2p, 3p, 2i, 3i, 2u, up, pi, ip, 2in, 3in, inp, pin, pni) that systematically cover the compositional combinations of these four operators, which are widely adopted in the CQA literature.
While traditional approaches treat CQA as an entity matching task on KGs, this work takes a different perspective by visualizing complex queries as images and challenging MLLMs to reason over them from the visual modality. As shown in Figure~\ref{figure::overview}, we define two tasks for MLLMs:
\begin{itemize}[leftmargin=8pt]
    \item \textbf{Complex Query Understanding (CQU):} recognizing its structural components of CVQ by telling a formatted expression.
    \item \textbf{Complex Query Reasoning (CQR):} performing multi-step logical reasoning over the query structure and predict the correct target entities $V_?$ with given reference subgraphs.
\end{itemize}

\section{Methodology}
To study \textbf{\data} (Visual \underline{\textbf{Qu}}ery \underline{\textbf{a}}s \underline{\textbf{S}}ymbolic \underline{\textbf{A}}nd \underline{\textbf{R}}easoning), we introduce the dataset construction pipeline and the two-stage training framework.

\subsection{Data Construction}
As shown in Figure~\ref{figure::overview}, the construction of \textbf{\data} consists of five steps: instance data sampling, semantic-based image selection, visual query synthesis, question and reference preparation, and CoT synthesis.

\paragraph{Step 0. Data Source.}
We employ three MMKGs as data sources: VisualSem~\cite{visualsem}, MKG-Y~\cite{MMRNS}, and FB15K-237~\cite{Freebase-SIGMOD08}, which collectively provide large-scale commonsense and encyclopedic knowledge spanning multiple data origins. Further details are provided in Appendix~\ref{appendix::data_source}.

\paragraph{Step 1. Instance Data Sampling.}
We adopt a template-guided, traversal-based sampling procedure following~\cite{BetaE}. For each predefined query template, we sample an answer entity and instantiate the query via backward traversal on the KG, recursively grounding predecessor entities and relations along inverse projection edges until all anchor nodes are identified. The instantiated query is then executed with the corresponding set operations (projection, intersection, union, and negation) to obtain its answer set. Duplicate queries and those with empty answer sets are discarded. This yields a query instance set $\mathcal{Q}_{i}=\{q_{i,1}, \ldots, q_{i,n(i)}\}$ for each query type $i$, where $n(i)$ denotes the number of sampled instances. The complete sampling procedure is detailed in Appendix~\ref{appendix::data_sampling}.

\paragraph{Step 2. Semantic-based Image Selection.}
To transfer query instances into the visual modality, we must obtain reliable multi-modal representations for each entity. While all three MMKGs provide image sets $\mathcal{I}(e)$ for each entity $e$, we observe that many entity images in the original MMKGs are noisy and semantically misaligned with their corresponding entity descriptions. To address this, we propose a semantic-based image selection strategy leveraging CLIP~\cite{CLIP} to measure the semantic consistency between each candidate image and its entity's textual description:
\begin{equation}
    I_{e}^{*}=\arg\max_{I\in\mathcal{I}(e)} \cos \left(\mathcal{M}_{\mathrm{vis}}(I),\ \mathcal{M}_{\mathrm{txt}}(T(e))\right)
\end{equation}
where $\mathcal{M}_{\mathrm{vis}}$ and $\mathcal{M}_{\mathrm{txt}}$ are the visual and textual encoders of CLIP, respectively, and $T(e)$ denotes the textual description of entity $e$. This strategy is motivated by the observation that textual entity information in MMKGs is substantially more accurate than the associated images, making text a reliable supervisory signal for image filtering. To further promote data diversity, we retain the top-3 images with the highest similarity scores for each entity, ensuring variety when the same entity appears across multiple query instances.

\paragraph{Step 3. Visual Query Synthesis.}
After obtaining semantically aligned entity images, we synthesize visual queries by organizing each sampled query $q$ into a directed computational graph and rendering it using GraphViz~\cite{graphviz}. Unlike existing works~\cite{m3str} that simply visualize multi-modal subgraphs sampled from KGs, we construct structured computational graphs that explicitly represent the logical composition of complex queries. Following the design language detailed in Appendix~\ref{appendix::query_visualization}, the four FOL operators are rendered with distinct visual encodings: intersection and union are represented as explicit operator nodes; relation projections correspond to directed, relation-labeled edges; and negation is encoded as a branch-level modifier, where the negated edge label is prefixed with \texttt{[NOT]} and highlighted in red to distinguish from positive ones.

\paragraph{Step 4. Question and Reference Preparation.}
We prepare task-specific text prompts for CQU and CQR. For CQU, a concise question template is used to instruct the model to interpret the given visual query. For CQR, we additionally provide a reference subgraph as reasoning context, within which the correct answers are embedded among distractor triples. Specifically, for each query, we extract core evidence triples that support valid reasoning paths to the answers, and supplement them with controlled distractor triples sampled from neighboring nodes to prevent trivial lookup. This design avoids the impracticality of ranking over the full KG while preserving meaningful reasoning complexity. The detailed subgraph sampling procedure is described in Appendix~\ref{appendix::reference_sample}.

\begin{figure}[t]
  \centering
  \includegraphics[width=0.5\textwidth]{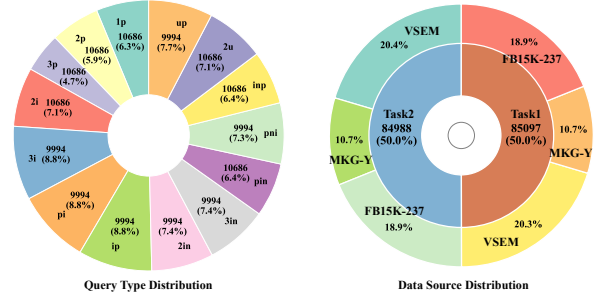}
  \caption{Overview of the query type and data source distribution of the {\data} dataset.}
  \label{figure::dataset}
  \vspace{-16pt}
\end{figure}
\begin{table*}[]
\caption{In-distribution experiment results of CQU/CQR on 14 different query types.}
\label{table::main_exp}
\vspace{-8pt}
\centering
\resizebox{0.9\textwidth}{!}{
\begin{tabular}{c|c|cccccccccccccc}
\toprule
\textbf{Model} & \textbf{Setting} & \textbf{1p} & \textbf{2i} & \textbf{2in} & \textbf{2p} & \textbf{2u} & \textbf{3i} & \textbf{3in} & \textbf{3p} & \textbf{inp} & \textbf{ip} & \textbf{pi} & \textbf{pin} & \textbf{pni} & \textbf{up} \\
\midrule
\multicolumn{16}{c}{\textbf{\textit{Task1: Complex Query Understanding (CQU)}}} \\
\midrule
\textbf{GPT-5.2} & \textbf{Zero-shot} & 88.40 & 83.60 & 87.71 & 78.51 & 85.88 & 88.61 & 79.93 & 74.70 & 60.68 & 69.26 & 29.97 & 26.23 & 7.72 & 54.40 \\
\textbf{Gemini-3-Flash} & \textbf{Zero-shot} & 93.16 & 98.06 & 97.72 & 86.75 & 96.55 & 94.05 & 93.80 & 88.89 & 95.69 & 88.39 & 92.20 & 92.26 & 83.60 & 86.64 \\
\midrule
\multirow{5}{*}{\textbf{Qwen3-VL-2B}} & \textbf{Zero-shot} & 0.00 & 4.59 & 0.15 & 0.00 & 0.00 & 1.14 & 0.00 & 0.00 & 0.00 & 0.00 & 0.00 & 0.00 & 0.00 & 0.00 \\
 & \textbf{SFT} & 96.87 & 97.79 & 97.88 & 98.09 & 98.69 & 98.16 & 98.05 & 98.70 & 98.20 & 98.68 & 98.32 & 98.30 & 96.95 & 98.70 \\
 & \textbf{DPO} & 97.63 & 98.14 & 98.17 & 96.58 & 98.85 & 97.47 & 96.25 & 96.33 & 98.38 & 98.22 & 98.19 & 97.64 & 97.51 & 98.62 \\
 & \textbf{ORPO} & 97.06 & 97.18 & 98.02 & 95.68 & 98.77 & 98.35 & 97.71 & 96.69 & 98.38 & 98.42 & 97.98 & 98.11 & 97.43 & 98.53 \\
 & \textbf{SimPO} & 97.82 & 98.24 & 98.33 & 96.78 & 98.77 & 96.77 & 84.99 & 97.40 & 98.56 & 98.42 & 97.65 & 97.65 & 97.43 & 98.77 \\
\midrule
\multirow{5}{*}{\textbf{Qwen3-VL-8B}} & \textbf{Zero-shot} & 14.83 & 0.53 & 0.91 & 62.05 & 0.16 & 0.38 & 0.33 & 0.47 & 0.00 & 0.00 & 14.25 & 57.17 & 0.32 & 0.00 \\
 & \textbf{SFT} & 98.19 & 99.39 & 99.39 & 99.60 & 99.17 & 99.43 & 97.88 & 98.93 & 99.10 & 98.94 & 98.72 & 99.25 & 98.55 & 99.35 \\
 & \textbf{DPO} & 98.48 & 99.38 & 99.70 & 99.50 & 99.17 & 99.30 & 97.88 & 99.06 & 99.10 & 98.94 & 98.86 & 99.25 & 98.96 & 99.10 \\
 & \textbf{ORPO} & 98.58 & 99.47 & 99.39 & 98.79 & 99.17 & 99.55 & 97.31 & 98.70 & 98.92 & 98.94 & 98.32 & 98.68 & 98.39 & 99.27 \\
 & \textbf{SimPO} & 98.29 & 99.20 & 99.39 & 99.30 & 99.51 & 99.24 & 97.88 & 99.06 & 99.10 & 98.94 & 98.46 & 99.06 & 98.96 & 99.10 \\
\midrule
\multicolumn{16}{c}{\textbf{\textit{Task2: Complex Query Reasoning (CQR)}}} \\
\midrule
\textbf{GPT-5.2} & \textbf{Zero-shot} & 100.00 & 99.47 & 88.62 & 69.48 & 72.25 & 99.62 & 90.86 & 45.63 & 22.98 & 65.30 & 85.48 & 35.85 & 65.11 & 58.47 \\
\textbf{Gemini-3-Flash} & \textbf{Zero-shot} & 99.43 & 100.00 & 100.00 & 100.00 & 96.39 & 100.00 & 100.00 & 99.53 & 99.64 & 99.87 & 99.46 & 99.25 & 96.30 & 100.00 \\
\midrule
\multirow{5}{*}{\textbf{Qwen3-VL-2B}} & \textbf{Zero-shot} & 0.19 & 0.88 & 3.03 & 0.60 & 1.31 & 0.76 & 1.79 & 0.00 & 0.54 & 0.53 & 1.08 & 0.19 & 0.80 & 0.65 \\
 & \textbf{SFT} & 100.00 & 99.82 & 100.00 & 99.30 & 100.00 & 100.00 & 100.00 & 98.42 & 98.93 & 99.93 & 99.86 & 99.26 & 99.52 & 99.11 \\
 & \textbf{DPO} & 100.00 & 99.82 & 100.00 & 99.10 & 100.00 & 100.00 & 100.00 & 98.42 & 99.02 & 99.93 & 99.86 & 99.35 & 99.44 & 98.95 \\
 & \textbf{ORPO} & 100.00 & 99.82 & 100.00 & 99.00 & 100.00 & 100.00 & 100.00 & 98.54 & 98.84 & 99.93 & 99.86 & 99.35 & 99.52 & 99.11 \\
 & \textbf{SimPO} & 100.00 & 99.82 & 100.00 & 99.30 & 100.00 & 100.00 & 100.00 & 98.42 & 98.93 & 99.93 & 99.86 & 99.35 & 99.52 & 98.95 \\
\midrule
\multirow{5}{*}{\textbf{Qwen3-VL-8B}} & \textbf{Zero-shot} & 47.72 & 37.04 & 18.66 & 8.43 & 11.00 & 45.95 & 29.69 & 4.96 & 4.49 & 3.69 & 14.38 & 6.79 & 11.41 & 3.09 \\
 & \textbf{SFT} & 100.00 & 100.00 & 100.00 & 99.60 & 100.00 & 100.00 & 99.91 & 99.75 & 99.91 & 100.00 & 99.93 & 99.81 & 99.84 & 99.60 \\
 & \textbf{DPO} & 100.00 & 100.00 & 100.00 & 99.80 & 100.00 & 100.00 & 99.91 & 99.64 & 99.64 & 100.00 & 99.93 & 99.81 & 99.68 & 99.68 \\
 & \textbf{ORPO} & 100.00 & 100.00 & 100.00 & 99.80 & 100.00 & 100.00 & 99.91 & 99.75 & 99.91 & 100.00 & 99.93 & 99.81 & 99.68 & 99.68 \\
 & \textbf{SimPO} & 100.00 & 100.00 & 100.00 & 99.60 & 100.00 & 100.00 & 99.91 & 99.64 & 99.73 & 100.00 & 99.93 & 99.81 & 99.84 & 99.60 \\
\bottomrule
\end{tabular}
}
\vspace{-16pt}
\end{table*}

\paragraph{Step 5. CoT Synthesis.}
Finally, we synthesize chain-of-thought (CoT) reasoning processes for each visual query instance. We design fine-grained, step-by-step CoT templates specifically tailored to each combination of query type and task type, yielding $14 \times 2 = 28$ templates in total. As shown in Figure~\ref{figure::overview}, the CoT consists of 4 steps for CQU and 7 steps for CQR, reflecting the greater reasoning complexity of the latter. Representative templates are illustrated in Appendix~\ref{appendix::prompt_template}. Upon completing the five steps above, the final dataset is formulated as $\mathcal{D}=\{(\mathcal{I}_i, \mathcal{Q}_i, \mathcal{A}_i)\}_{i=1}^N$, where $\mathcal{I}_i$ denotes the visual query image, $\mathcal{Q}_i$ the question prompt (with reference subgraph for CQR), and $\mathcal{A}_i$ the CoT annotation with final answers.

\subsection{Dataset Overview}
As shown in Figure~\ref{figure::dataset}, \textbf{\data} covers all 14 query types with a balanced type distribution across three MMKG sources. We ensure approximate parity in data volume across query types, and maintain a near 1:1 ratio between CQU and CQR instances. The dataset is partitioned into training, validation, and test sets following an 8:1:1 split ratio.

\subsection{Training and Evaluation Protocol}
Our preliminary experiments reveal that current MLLMs exhibit poor zero-shot performance on CVQ tasks, which we attribute to two compounding factors. First, CVQ is a novel data type absent from standard pre-training corpora, leaving MLLMs without a foundational schema for processing such inputs. Second, complex query patterns demand fine-grained structural recognition and multi-step logical reasoning that exceed the capabilities of general-purpose MLLMs. These observations motivate a two-stage training framework designed to progressively build CVQ competence.

\paragraph{Stage 1: Supervised Fine-tuning (SFT).}
We first apply SFT on our large-scale synthetic \textbf{\data} dataset, enabling MLLMs to internalize the visual query processing patterns encoded in our CoT templates. This stage establishes a foundational ability to parse CVQ structures and follow structured CoT for both CQU and CQR tasks.

\paragraph{Stage 2: Fine-grained Preference Optimization.}
While SFT instills basic CVQ competence, MLLMs still struggle with fine-grained structural recognition, particularly in distinguishing subtle differences among query components. To address this bottleneck, we collect hard samples on which the SFT model underperforms, pair them with model-generated negative outputs to construct preference data, and apply preference alignment training to sharpen MLLMs' capabilities. We experiment with popular alignment objectives including DPO~\cite{DPO}, ORPO~\cite{ORPO}, and SimPO~\cite{SimPO}.
\paragraph{Evaluation.} We evaluate CQU/CQR with accuracy metrics. For CQU, model responses are structuredly parsed and evaluated based on logical equivalence with the ground-truth query description. For CQR, a prediction is considered correct if and only if the predicted entity set exactly matches the ground-truth answers, regardless of order.

\section{Experiments and Evaluation}
We present comprehensive experiments on \textbf{\data} to evaluate MLLMs across three dimensions: in-distribution CVQ performance, out-of-distribution generalization within CVQ tasks, and transferability of CVQ-trained models to non-CVQ visual reasoning tasks.

\subsection{Experiment Settings}
We adopt Qwen3-VL-2B/8B~\cite{Qwen3-VL} as our backbone models. We also experimented with the LLaVA~\cite{llava-1.5} series, but found that \textbf{\data} instances exceed LLaVA's maximum context length and are thus excluded. All models are trained on 8$\times$A100 GPUs with LoRA~\cite{LoRA-ICLR22}. Detailed hyper-parameter configurations are provided in Appendix~\ref{appendix::implementation_details}.

\subsection{In-Distribution Experiments}
We present the main experimental results in Table~\ref{table::main_exp}. Several key observations emerge from these results.

\par \textbf{(1). Zero-shot performance of MLLMs on CVQ is severely limited.} Small-size open-source MLLMs exhibit near-zero performance on both CQU and CQR tasks in the zero-shot setting, indicating that CVQ represents a fundamentally novel reasoning paradigm absent from standard pre-training. After two-stage training, however, the same models acquire substantial CVQ competence, demonstrating that the primary bottleneck lies in the lack of exposure to FOL-compositional visual reasoning patterns rather than inherent capacity.

\par \textbf{(2).Training consistently outperforms proprietary models.} Despite their strong general capabilities, proprietary MLLMs such as GPT and Gemini underperform compared to smaller but fine-tuned open-source models. This highlights a critical limitation of closed-source systems: the inability to adapt to novel, structured reasoning tasks through task-specific training.

\par \textbf{(3). Two-stage training provides additional but moderate gains.} The preference optimization stage yields consistent improvements over SFT alone, though the margins are not dramatic. We attribute this to rapid convergence during SFT on our large-scale dataset, which brings the model close to its in-distribution performance ceiling. We emphasize that the primary goal of this experiment is not to achieve SOTA performance, but to validate that MLLMs can acquire in-distribution CVQ generalization. The specific advantages of the second training stage are further examined.

\subsection{Scalability Experiments}
\begin{figure}[t]
  \centering
  \vspace{-24pt}
  \includegraphics[width=0.5\textwidth]{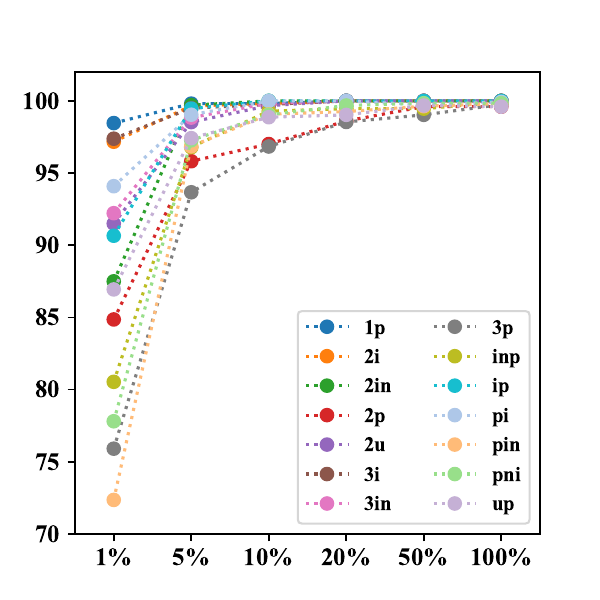}
  \caption{Scalability experiment results with different data proportions on 14 query types with the CQR task.}
  \label{figure::scala}
  \vspace{-16pt}
\end{figure}
Given that MLLMs achieve near-perfect in-distribution performance when trained on the full {\data} dataset, a natural question arises: to what extent does this generalization depend on data scale? We investigate this through scalability experiments on {\data}.
As shown in Figure~\ref{figure::scala}, we report CQR performance of Qwen3-VL-8B trained on 1\% to 100\% of {\data} across all 14 query types. Two key findings emerge:

\noindent\textbf{(1) Overall convergence is achieved with 20\%--50\% of the data.} MLLMs reach near-perfect performance after training on approximately 20\% to 50\% of the full dataset, with additional data yielding only marginal gains. This suggests that \textbf{\data} provides sufficient coverage of FOL reasoning patterns within a moderate data scale.

\noindent\textbf{(2) Query complexity governs convergence rate.} Different query types exhibit markedly different convergence behaviors, with more complex queries consistently requiring more training data. For instance, the simplest 1p query achieves near-full performance with as little as 1\% of the data, whereas compositionally complex queries involving three logical operators, such as \texttt{pin}, \texttt{pni}, and \texttt{inp}, require the full dataset to reach stable convergence.

These results reveal a clear correlation between query complexity and data requirement: simple query patterns can be mastered with limited supervision, while complex compositional patterns necessitate broader data coverage to achieve reliable generalization.

\subsection{Extrapolation Experiments}
We further examine the extrapolation properties of {\data} from two perspectives: generalization to larger references and across different query types.
\subsubsection{Extrapolation with Larger Contexts}
\begin{figure}[t]
  \centering
  \includegraphics[width=0.5\textwidth]{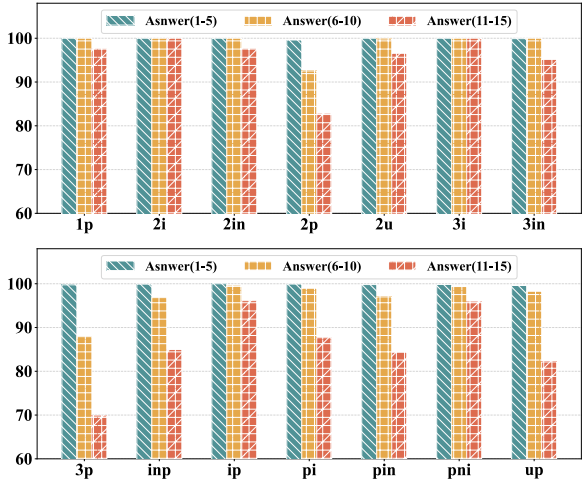}
  \caption{Accuracy of Qwen3-VL-8B for larger context extrapolation experiments on 14 query types.}
  \label{figure::large_contexts_transfer}
  \vspace{-16pt}
\end{figure}
\begin{figure*}[h]
  \centering
  \includegraphics[width=\textwidth]{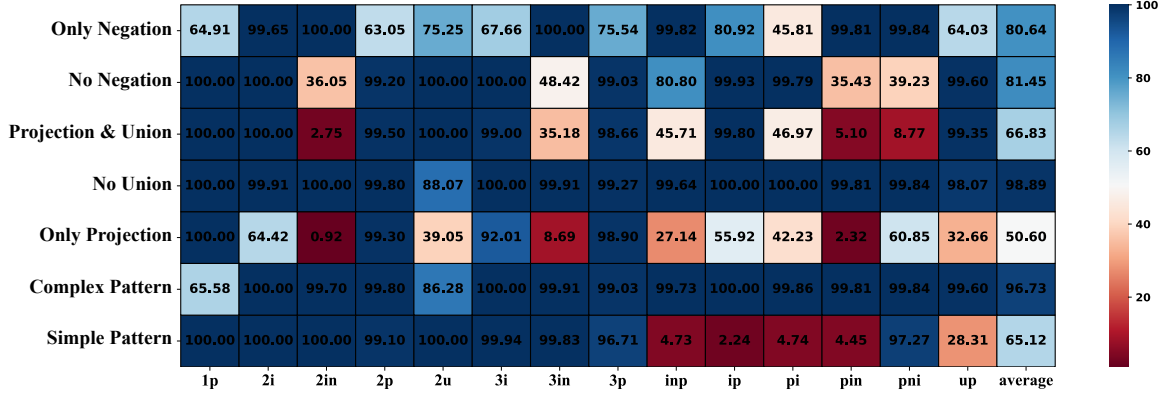}
  \vspace{-16pt}
  \caption{Experiment results for extrapolation across different query types. We designed seven experimental groups, each containing only a specific type set. We train MLLMs on these groupsnand infer on all query types.}
  \label{figure::query_type_transfer}
\end{figure*}
In the original dataset, reference subgraphs are constructed with at most 5 answers per query. To assess whether models trained on this setting can generalize to larger reference contexts, we additionally sample non-overlapping test instances with answer counts of 6--10 and 11--15, constructing correspondingly larger reference subgraphs for inference.

As shown in Figure~\ref{figure::large_contexts_transfer}, model accuracy degrades notably as reference context size increases, with the degree of degradation varying across query types. Two consistent patterns emerge:

\noindent\textbf{(1). Query complexity amplifies extrapolation difficulty.} Simple query types such as \texttt{1p}, \texttt{2i}, and \texttt{2in} maintain near-perfect accuracy even under larger contexts, whereas compositionally complex queries such as \texttt{3p}, \texttt{pin} exhibit significant performance drops. This suggests that complex reasoning structures are more sensitive to context scaling.

\noindent\textbf{(2). Relation projection is more context-sensitive than logical operators.} Queries dominated by projection operations (\texttt{1p}, \texttt{2p}, \texttt{3p}) suffer disproportionately larger accuracy drops compared to queries with the same number of hops but involving union, intersection, or negation. This indicates that navigating relational chains over expanded reference graphs poses a greater challenge for MLLMs than applying logical operators.

\subsubsection{Extrapolation across Query Types}
We investigate whether MLLMs trained on a subset of query types can generalize to unseen types. We design 7 training groups with distinct compositional configurations: (1) simple patterns only, (2) complex patterns only, (3) projection only, (4) no union, (5) projection and union only (no intersection or negation), (6) no negation, and (7) negation only. Detailed split configurations are provided in Table~\ref{table::query_type_data_split}. The cross-type generalization results are visualized as a heatmap in Figure~\ref{figure::query_type_transfer}. Three key findings emerge:

\noindent\textbf{(1) Complex patterns subsume simple ones, but not vice versa.} Models trained on complex query patterns generalize well to simpler types, whereas models trained exclusively on simple patterns fail to generalize to complex ones. This asymmetry is consistent with the compositional nature of complex queries: since complex patterns are built upon combinations of simpler operators, training on them implicitly covers the simpler cases. The reverse, however, does not hold, as compositional combinations introduce reasoning structures that cannot be extrapolated from simple patterns alone.

\noindent\textbf{(2) Projection alone fails to generalize to union and intersection.} As revealed by comparing groups (3)--(5), projection-only training does not transfer to union or intersection query types. Conversely, training on combinations that include union enables some generalization to union-involving queries, but intersection remains difficult to acquire through training with other operators, suggesting it requires dedicated supervision.

\noindent\textbf{(3) Negation is the hardest operator to transfer.} Models trained without negation fail to handle negation-involving queries, and even joint training with other operators (p/u/i) does not yield reliable negation generalization. Taken together, these findings reveal a clear operator hierarchy in terms of transferability: union is relatively easy to generalize across training configurations, while projection, intersection, and negation each require explicit training coverage. This highlights the importance of query type diversity in CVQ training and exposes fundamental limitations in the logical reasoning capabilities of current MLLMs.

\subsection{General Capability Experiments}
\begin{figure}[t]
  \centering
  \includegraphics[width=0.5\textwidth]{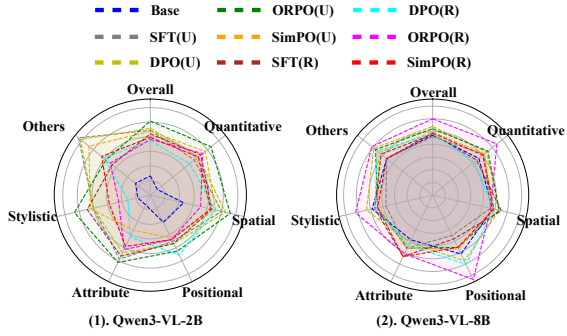}
  \caption{Evaluation results on VisuLogic after training on our {\data} data.}
  \label{figure::visuallogix}
  \vspace{-8pt}
\end{figure}
To investigate the transferability of \textbf{\data} training to broader abstract reasoning tasks, we evaluate our trained MLLMs on VisuLogic~\cite{visulogic}, a benchmark covering diverse abstract visual reasoning tasks including graphical reasoning. As shown in Figure~\ref{figure::visuallogix}, two key findings emerge from the 7-subtask evaluation:

\noindent\textbf{(1). CVQ training stimulates latent abstract reasoning capabilities.} Both Qwen3-VL-2B and 8B models achieve consistent improvements across all VisuLogic subtasks after \textbf{\data} training, with the 2B model showing particularly pronounced gains, suggesting CVQ training is especially effective at unlocking abstract reasoning potential in smaller models.

\noindent\textbf{(2). Two-stage training yields superior OOD generalization.} Models trained with preference alignment (DPO, ORPO, SimPO) consistently outperform SFT-only models on VisuLogic, validating the unique value of our two-stage design for transferable reasoning.

\par We further evaluate the impact of \textbf{\data} training on general-purpose benchmarks, including AI2D~\cite{AI2D}, MathVerse~\cite{MathVerse}, OCRBench~\cite{OCRBench}, and TextVQA~\cite{TextVQA}. As shown in Figure~\ref{table::common}, {\data} training does not induce catastrophic forgetting: the model retains strong general multi-modal capabilities and even achieves improvements on MathVerse and AI2D, with CQR-trained models showing larger gains on the latter. Minor fluctuations are observed on OCRBench and TextVQA. These results confirm that our synthetic training data enhances abstract reasoning without compromising general-purpose performance.
\begin{table}[t]
\caption{General benchmark performance after training.}
\label{table::common}
\centering
\resizebox{0.45\textwidth}{!}{
\begin{tabular}{cc|cccc}
\toprule
\multicolumn{2}{c|}{\textbf{Setting}} & \textbf{AI2D} & \textbf{MATH} & \textbf{OCR} & \textbf{VQA} \\
\midrule
\multicolumn{2}{c|}{Qwen3-VL-8B} & 40.45 & 50.56 & 80.80 & 93.26 \\
\midrule
\multirow{4}{*}{CQU} & SFT & 33.32 & 50.00 & 78.70 & 92.06 \\
 & DPO & 34.16 & 51.85 & 78.20 & 91.92 \\
 & ORPO & 33.26 & 51.85 & 79.00 & 91.88 \\
 & SimPO & 34.00 & 52.04 & 78.40 & 92.00 \\
\midrule
\multirow{4}{*}{CQR} & SFT & 38.96 & 52.04 & 77.40 & 92.34 \\
 & DPO & 43.30 & 54.44 & 80.50 & 92.28 \\
 & ORPO & 43.30 & 54.44 & 80.80 & 92.44 \\
 & SimPO & 43.33 & 54.63 & 80.60 & 92.38 \\
\bottomrule
\end{tabular}
}
\vspace{-8pt}
\end{table}

\section{Conclusion}
In this paper, we introduce complex visual queries (CVQs) as a novel research direction for symbolic and abstractive reasoning in MLLMs. We formalize two tasks, CQU and CQR, grounded in first-order logic operators over MMKGs, and construct \textbf{\data}, a large-scale benchmark covering 14 canonical FOL query types with fine-grained CoT annotations. We further propose a two-stage training framework combining SFT with preference alignment to progressively build CVQ competence.
Our experiments reveal several key findings: CVQ generalization can be effectively acquired through training; convergence difficulty scales with query complexity; cross-type transfer is asymmetric across FOL operators; and CVQ training consistently benefits OOD abstract reasoning, with two-stage training proving especially advantageous.
In the future, we plan to extend CVQ toward agentic workflows for neuro-symbolic visual reasoning over dynamic knowledge sources. We hope this work serves as a solid foundation for research at the intersection of symbolic reasoning, knowledge graphs, and multi-modal understanding.

\section*{Limitations}
Despite the substantial work and technical contributions made in this paper, it still has several limitations, which are summarized as follows:
\paragraph{Diversity of the {\data} data.} Our data primarily comes from three general-purpose knowledge graphs, and we lack data from specific domains. Currently, our research focuses mainly on general-purpose domains, with little consideration for various knowledge-intensive vertical fields.

\paragraph{Limitation of the two CVQ tasks.} Compared to the original CQA tasks, the two tasks we have designed so far have been simplified to some extent for the MLLM setting and therefore have certain limitations. In the future, we will attempt to build an agentic workflow that allows MLLM to go beyond simply answering given questions and context, enabling it to actively explore the MMKG and provide global answers.

\paragraph{Lack of further exploration for CVQ.} The problem of CQA in real-world scenarios is a complex reasoning problem with a much larger search space. In this work, we have only made a preliminary start on this topic and have not conducted a more in-depth investigation. Ultimately, the models trained on Task 1 and Task 2 should be deployed as agents to enable automated complex reasoning.

\section*{Ethics Statement}
In this paper, we utilize open-source MMKGs as our data sources to build datasets. Additionally, the primary MLLM backbones we employ are mainstream open-source models. We did not collect data or conduct computational experiments in ways that violated scientific ethics. Therefore, our work does not involve any ethical issues.


\bibliography{custom}

\appendix

\section{Dataset Information}
\begin{table*}[h]
\caption{Statistical information about the MMKG data source used in our data engine.}
\label{table::data_source}
\centering
\begin{tabular}{c|cccc}
\toprule
\textbf{Dataset} & \textbf{Entity} & \textbf{Relation} & \textbf{Triple} & \textbf{Data Source} \\
\midrule
\textbf{FB15K-237} & 14541 & 237 & 310116 & FreeBase \\
\textbf{MKG-Y} & 15000 & 16 & 26638 & YAGO \\
\textbf{VisualSem} & 89896 & 13 & 1481007 & Wikipedia, ImageNet, BabelNet \\
\bottomrule
\end{tabular}
\end{table*}
\subsection{Details of the Data Source}
\label{appendix::data_source}
We present the detailed information of the MMKGs used in our data engine in Table \ref{table::data_source}. These three datasets have different data sources. FB15K-237 is from FreeBase, MKG-Y is built on YAGO \cite{Yago-WWW07}, and VisualSem is constructed based on Wikipedia \cite{DBLP:journals/cacm/wikidata}, ImageNet \cite{imagenet}, and BabelNet \cite{babelnet}.

\subsection{Details of Dataset Construction}
\subsubsection{Data Instance Sampling}
\label{appendix::data_sampling}
We adopt the template-based query sampling procedure from BetaE~\cite{BetaE}. For each predefined query template, we sample an answer entity and instantiate a symbolic query by backward traversal on the knowledge graph, where predecessor entities and relations are recursively sampled along the inverse direction of projection edges until all anchor nodes are grounded. The instantiated query is then executed with the corresponding set operations, including projection, intersection, union, and complement, to obtain its answer set.

We discard duplicate queries and queries with empty answer sets. A loose upper bound on the answer-set size is used only to remove pathological cases with excessively large answers, rather than to strictly constrain the number of answers. Finally, entity and relation identifiers are mapped to textual labels to verbalize the sampled instances. The overall sampling procedure is summarized in Algorithm~\ref{alg:query_sampling}.

\begin{algorithm}[h]
\caption{Template-guided Complex Query Sampling}
\label{alg:query_sampling}
\begin{algorithmic}[1]
\REQUIRE Knowledge graph $\mathcal{G}=(\mathcal{E}, \mathcal{R}, \mathcal{T})$, query templates $\mathcal{S}$, per-template query budget $N$, answer-size threshold $M$
\ENSURE Sampled query-answer pairs $\mathcal{D}$

\STATE Build forward and backward adjacency indices for $\mathcal{G}$
\STATE $\mathcal{D} \leftarrow \emptyset$

\FOR{each template $s \in \mathcal{S}$}
    \STATE $\mathcal{D}_s \leftarrow \emptyset$
    \STATE $\mathcal{Q}_s \leftarrow \emptyset$
    \WHILE{$|\mathcal{D}_s| < N$}
        \STATE Sample an answer entity $a \in \mathcal{E}$
        \STATE Instantiate $s$ by backward traversal from $a$ to obtain query $q$
        \STATE Execute $q$ on $\mathcal{G}$ to obtain answer set $A(q)$
        \IF{$q \notin \mathcal{Q}_s$ \AND $0 < |A(q)| \leq M$}
            \STATE $\mathcal{D}_s \leftarrow \mathcal{D}_s \cup \{(q, A(q))\}$
            \STATE $\mathcal{Q}_s \leftarrow \mathcal{Q}_s \cup \{q\}$
        \ENDIF
    \ENDWHILE
    \STATE $\mathcal{D} \leftarrow \mathcal{D} \cup \mathcal{D}_s$
\ENDFOR

\STATE Verbalize queries and answers by mapping entity and relation identifiers to textual labels
\RETURN $\mathcal{D}$
\end{algorithmic}
\end{algorithm}

\subsubsection{Design Language for Visualized Query}
\label{appendix::query_visualization}

We convert each symbolic query into a directed visual graph, where nodes represent anchor entities, latent entity sets, answer sets, or logical operators, and edges represent relational projections. The graph is rendered from left to right to preserve the compositional order of the original query.

Anchor entities are visualized with representative images selected from pre-computed vision-language matching candidates, while intermediate and answer variables are shown as masked set nodes, denoted by ``Entity Set A'', ``Entity Set B'', etc. These masked nodes expose the reasoning structure without revealing the ground-truth entities.

A relational projection is represented by a directed edge labeled with the corresponding relation. Thus, path queries such as 1p, 2p, and 3p are visualized as chains of relation-labeled edges. Conjunctive queries use an explicit intersection operator node, where multiple incoming branches are merged and connected to the resulting set. Disjunctive queries use a union operator node to merge alternative branches.

Negation is encoded as a branch-level modifier: a negated relation is marked by prefixing its edge label with [NOT], and the corresponding edge is highlighted in red to distinguish negative constraints from positive ones. This design supports negative query templates such as 2in, 3in, pin, pni, and inp.

\subsubsection{Reference Subgraph Sampling}
\label{appendix::reference_sample}

For each Task2 query, we construct a compact reference subgraph as its reasoning context. Let $\mathcal{A}^{\mathrm{full}}$ denote the original answer set. Given an answer-budget range $(A_{\min}, A_{\max})$, we discard queries with fewer than $A_{\min}$ original answers. For each remaining query, we first sample an answer budget
\[
k \sim \mathrm{Uniform}\left(\{A_{\min}, \ldots, \min(A_{\max}, |\mathcal{A}^{\mathrm{full}}|)\}\right),
\]
and then uniformly sample $k$ answers without replacement from $\mathcal{A}^{\mathrm{full}}$ to form the target answer set $\mathcal{A}$.

For queries involving negation, let $\mathcal{N}^{\mathrm{full}}$ denote the set of non-result entities used to ground the negated branch. We uniformly sample at most $N_{\max}$ entities from $\mathcal{N}^{\mathrm{full}}$ to obtain $\mathcal{N}$, which is used only for constructing evidence of the negated branch.

We extract core evidence triples according to the logical structure of each query. For projection queries, we retain only reasoning paths that reach the sampled answers in $\mathcal{A}$. For intersection and union queries, we retain supporting triples from branches that contribute to at least one sampled answer. For negation queries, positive branches are grounded by sampled answers, while negated branches are grounded by sampled non-result entities in $\mathcal{N}$. For composite queries, an intermediate entity is retained only if it participates in at least one valid reasoning chain leading to a sampled answer or grounding a required negated branch. The resulting set of core evidence triples is denoted by $\mathcal{E}_c$.

To avoid providing a context that consists only of gold reasoning paths, we add controlled distractor triples. Let $\mathcal{V}_c$ denote the set of nodes expanded for distractor sampling, consisting of source entities appearing in $\mathcal{E}_c$ and the sampled answer entities in $\mathcal{A}$. For each $v \in \mathcal{V}_c$, we sample at most $M_{\max}$ outgoing triples $(v,r,u)$ such that $(v,r)$ does not appear in $\mathcal{E}_c$ and $u \notin \mathcal{V}_c$. This prevents distractors from duplicating core query patterns or directly connecting to existing core nodes. Let $\mathcal{E}_d$ denote the sampled distractor triples. The final reference context is
\[
\mathcal{C} = \mathcal{E}_c \cup \mathcal{E}_d .
\]

All random choices are made with a fixed pseudo-random seed and a deterministic query processing order for reproducibility. We instantiate this procedure under three answer-budget settings: $1$--$5$, $6$--$10$, and $11$--$15$, corresponding to $(A_{\min}, A_{\max})=(1,5)$, $(6,10)$, and $(11,15)$, respectively. For all three settings, we use $N_{\max}=8$ and $M_{\max}=4$.

\subsubsection{CoT Templates}
\label{appendix::prompt_template}
In this section, we present all the CoT templates used for 14 different query types and 2 task types in Figure~\ref{box:cot_task1} and Figure~\ref{box:cot_task2}. The complete prompt templates are uniquely designed for each query type and task type, resulting in a total of 28 templates. Given this large number, we present only one example for each task type in the paper; the complete templates are available in our supplementary materials.

\subsubsection{Dataset Statistics}
\begin{table*}[]
\caption{The detailed statistics for CQU/CQR of {\data} data.}
\label{table::full_dataset}
\centering
\begin{tabular}{c|cccc|cccc}
\toprule
\multirow{2}{*}{\textbf{Query Type}} & \multicolumn{4}{c|}{\textbf{Task1}} & \multicolumn{4}{c}{\textbf{Task2}} \\
 & \textbf{Total} & \textbf{Train} & \textbf{Valid} & \textbf{Test} & \textbf{Total} & \textbf{Train} & \textbf{Valid} & \textbf{Test} \\
\midrule
\textbf{1p} & 5343 & 4298 & 519 & 526 & 5343 & 4301 & 522 & 520 \\
\textbf{2p} & 5000 & 4002 & 500 & 498 & 4994 & 4004 & 488 & 502 \\
\textbf{3p} & 4020 & 3163 & 434 & 423 & 3980 & 3153 & 416 & 411 \\
\textbf{2i} & 6000 & 4794 & 639 & 567 & 6000 & 4798 & 637 & 565 \\
\textbf{3i} & 7500 & 5951 & 759 & 790 & 7500 & 5944 & 755 & 801 \\
\textbf{pi} & 7500 & 5984 & 772 & 744 & 7487 & 5990 & 770 & 727 \\
\textbf{ip} & 7500 & 6026 & 716 & 758 & 7499 & 6013 & 726 & 760 \\
\textbf{2in} & 6300 & 5019 & 622 & 659 & 6300 & 5013 & 631 & 656 \\
\textbf{3in} & 6300 & 5067 & 620 & 613 & 6300 & 5072 & 624 & 604 \\
\textbf{pin} & 5464 & 4397 & 537 & 530 & 5463 & 4392 & 532 & 539 \\
\textbf{pni} & 6200 & 4972 & 606 & 622 & 6186 & 4971 & 593 & 622 \\
\textbf{inp} & 5448 & 4335 & 556 & 557 & 5426 & 4304 & 562 & 560 \\
\textbf{2u} & 6000 & 4793 & 598 & 609 & 6000 & 4787 & 601 & 612 \\
\textbf{up} & 6522 & 5276 & 632 & 614 & 6510 & 5248 & 642 & 620 \\
\textbf{All} & 85097 & 68077 & 8510 & 8510 & 84988 & 67990 & 8499 & 8499 \\
\bottomrule
\end{tabular}
\end{table*}
We present the original statistical information of the datasets in Table~\ref{table::full_dataset}.
\section{Experiments}
\subsection{Implementation Details}
\label{appendix::implementation_details}

We implement all training experiments with LlamaFactory using PyTorch 2.6.0+cu124 and DeepSpeed 0.18.4. All experiments are conducted on 8 NVIDIA A100-SXM4-80GB GPUs. Unless otherwise specified, we train models with bf16 precision, LoRA rank 32, and a maximum sequence length of 32,768 tokens.

For supervised fine-tuning (SFT) on Task1, we train the model for 3 epochs with two learning rates, $3\times10^{-4}$ and $5\times10^{-4}$. For each learning rate, the per-device batch size is set to 4, with 32 gradient accumulation steps, resulting in an effective global batch size of 1024 over 8 GPUs. We evaluate both resulting checkpoints and report the average performance across the two learning-rate settings.

For SFT on CCQ, we use the same two learning rates and train for 3 epochs.
For each learning rate, the per-device batch size is set to 1, with 8 gradient accumulation steps, resulting in an effective global batch size of 64. Similarly, we evaluate both resulting checkpoints and report the average performance across the two learning-rate settings.

For the preference optimization stage, including DPO, ORPO, and SimPO, we initialize from the SFT checkpoint trained with a learning rate $3\times10^{-4}$ and keep the training setup unchanged, except for the optimization objective. Specifically, we train for 1 epoch with a learning rate of $1\times10^{-6}$, a per-device batch size of 1, and 1 gradient accumulation step, yielding an effective global batch size of 8. We set the preference optimization coefficient \texttt{pref\_beta} to 0.1.
\begin{table}[]
\caption{Detailed setting for 7 groups in the query type extrapolation experiments.}
\label{table::query_type_data_split}
\centering
\resizebox{0.5\textwidth}{!}{
\begin{tabular}{l|l}
\toprule
\multicolumn{1}{c|}{Group Setting} & \multicolumn{1}{c}{Query Type} \\ \midrule
(1). Simple Pattern & \texttt{1p}, \texttt{2p}, \texttt{2i}, \texttt{2in}, \texttt{2u} \\
(2). Complex Pattern & \texttt{3p}, \texttt{3i}, \texttt{pi}, \texttt{ip}, \texttt{3in}, \texttt{pin}, \texttt{pni}, \texttt{inp}, \texttt{up} \\
(3). Only Projection & \texttt{1p}, \texttt{2p}, \texttt{3p} \\
(4). No Union & \texttt{1p}, \texttt{2p}, \texttt{3p}, \texttt{2i}, \texttt{3i}, \texttt{pi}, \texttt{ip}, \texttt{2in}, \texttt{3in}, \texttt{pni}, \texttt{pin}, \texttt{inp} \\
(5). Projection \& Union & \texttt{1p}, \texttt{2p}, \texttt{3p}, \texttt{2u}, \texttt{up} \\
(6). No Negation & \texttt{1p}, \texttt{2p}, \texttt{3p}, \texttt{2i}, \texttt{3i}, \texttt{pi}, \texttt{ip}, \texttt{2u}, \texttt{up} \\
(7). Only Negation & \texttt{2in}, \texttt{3in}, \texttt{pin}, \texttt{pni}, \texttt{inp} \\ \bottomrule
\end{tabular}
}
\end{table}
\subsection{Extrapolation across Query Types}
Here we present the detailed settings for the 7 groups of {\data} data we used in this experiment.

\section{The Use of Large Language Models}
The primary research subject of this paper is LLM \& MLLM. Additionally, LLMs are employed \textbf{as a general assistant} for code debugging and polishing certain paragraphs. Core idea conception, experimental design, and paper writing are completed by human authors.

\clearpage
\begin{figure*}[h]
\centering
\begin{tcolorbox}[colback=gray!10!white, colframe=gray!60!black, title=CoT Template for Task1 CQU]
\begin{lstlisting}[
  basicstyle=\ttfamily\small,
  breaklines=true
]
[pin]
<think>
Step 1: Visual Entity Extraction
I identify the nodes in the image:
- Source Node 1: [{head1}]
- Intermediate Node: [Entity Set A]
- Source Node 2: [{head2}]
- Operation Symbol: [Intersection]
- Target Node: [Entity Set B]

Step 2: Structural Relationship Analysis
Next, I analyze the connections between the nodes:
The structure shows a multi-hop positive branch and a single-hop negative branch converging:
1. Arrow from "{head1}" to "Entity Set A" with label "'{relation1}'".
2. Dashed arrow from "Entity Set A" to the "Intersection" symbol with label "'{relation2}'".
3. A RED dashed arrow (labeled [NOT]) from "{head2}" to the "Intersection" symbol with label "'{relation3}'".
4. A dashed arrow points from the "Intersection" symbol to "Entity Set B".

Step 3: Semantic Triplet Formulation
Based on the visual components, I formulate the semantic triplets:
We are looking for a target entity set [?y] (Entity Set B) that satisfies a 2-hop positive path while excluding a negative condition.
Positive Condition (Path):
- Triplet 1: Subject [{head1}] -- Predicate ['{relation1}'] --> Object [?x] (Entity Set A)
- Triplet 2: Subject [?x] -- Predicate ['{relation2}'] --> Object [?y]
Negative Condition ([NOT]):
- Triplet 3: Subject [{head2}] -- Predicate ['{relation3}'] --> Object [?y]
The Intersection symbol with the [NOT] indicator implies: Find [?y] that results from the Path BUT DOES NOT exist in the Negative Triplet.

Step 4: Compositional Logic Inference
Determining the logical structure:
This is a Set Difference (Path Intersection with Negation) (pin) query.
Logic Chain:
1. Projection 1: Find intermediate entities [Entity Set A] connected to [{head1}] via ['{relation1}'].
2. Projection 2: Find potential targets connected to [Entity Set A] via ['{relation2}'].
3. Negative Search: Find entities connected to [{head2}] via ['{relation3}'].
4. Difference Operation: Subtract the results of the Negative Search from the results of Projection 2.
5. Result: Entity Set B.
</think>
<Answer>
(({head1}, '{relation1}', ?), '{relation2}', ?) AND (NOT ({head2}, '{relation3}', ?))
</Answer>
\end{lstlisting}
\end{tcolorbox}
\caption{The CoT prompt template for Task1 CQU. We only present pin query for demonstration due to the huge volume of 14 full templates. We submit the full templates in the supplemental materials.}
\label{box:cot_task1}
\end{figure*}

\clearpage
\begin{figure*}[h]
\centering
\begin{tcolorbox}[colback=gray!10!white, colframe=gray!60!black, title=CoT Template for Task2 CQR]
\begin{lstlisting}[
  basicstyle=\ttfamily\small,
  breaklines=true
]
[pin]
<think>
{task1_cot}

Step 5: Knowledge Context Analysis
Total knowledge triples available: {context_count}
I will filter each branch independently, then compute their difference.

Step 6: Step-by-step Filtering and Reasoning

=== PATH HOP 1: ({head1}, '{relation1}', ?) ===
Step 6.1.1: Retrieved {path_hop1_triple_count} triples for [{head1}]
{path_hop1_triples}
Step 6.1.2: Filtered by ['{relation1}']: {path_hop1_filtered_count} triples
{path_hop1_filtered_triples}
Intermediate entities: {intermediate_entities}

=== PATH HOP 2: (?x, '{relation2}', ?) ===
Step 6.2.1: For each intermediate entity, retrieve all triples
{path_hop2_candidate_triples_by_entity}
Step 6.2.2: Filtered by ['{relation2}']: {path_hop2_filtered_count} triples
{path_hop2_filtered_triples}
Path targets: {path_targets}

=== NEGATIVE BRANCH: ({head2}, '{relation3}', ?) ===
Step 6.3.1: Retrieved {negative_triple_count} triples for [{head2}]
{negative_triples}
Step 6.3.2: Filtered by ['{relation3}']: {negative_filtered_count} triples
{negative_filtered_triples}
Negative Branch result set: {negative_set}

=== SET DIFFERENCE ===
Step 6.4: Computing Path targets - Negative Branch result set
  |Path targets| = {path_targets}
  |Negative Branch result set| = {negative_set}
  |Result| = {result}

Step 7: Answer Derivation
Final Answer: The entities in the final result set are:
{answer_bullets}
</think>
<Answer>
{final_answer}
</Answer>
\end{lstlisting}
\end{tcolorbox}
\caption{The CoT prompt template for Task2 CQR. We only present pin query for demonstration due to the huge volume of 14 full templates. We submit the full templates in the supplemental materials.}
\label{box:cot_task2}
\end{figure*}

\end{document}